\documentclass[sigconf]{acmart}

\AtBeginDocument{%
  }

\copyrightyear{2025} 
\acmYear{2025} 
\setcopyright{acmlicensed}\acmConference[WWW '25]{Proceedings of the ACM Web Conference 2025}{April 28-May 2, 2025}{Sydney, NSW, Australia}
\acmBooktitle{Proceedings of the ACM Web Conference 2025 (WWW '25), April 28-May 2, 2025, Sydney, NSW, Australia}
\acmDOI{10.1145/3696410.3714612}
\acmISBN{979-8-4007-1274-6/25/04}

\usepackage{algorithm}
\usepackage{algorithmicx}
\usepackage{xspace}
\usepackage[commentColor=blue,rightComments=true,noEnd=false]{algpseudocodex}

\newcommand{\etal}{\emph{et al.}\xspace}

\usepackage{amssymb}
\usepackage{amsmath}
\usepackage{booktabs}
\usepackage{makecell}

\begin{document}

\title{FedRIR: Rethinking Information Representation in Federated Learning}

\author{Yongqiang Huang}
\orcid{0009-0008-2799-4726}
\affiliation{%
  \institution{Sichuan University}
  \department{School of Cyber Science and Engineering}
  \city{Chengdu}
  \state{Sichuan}
  \country{China}
}
\email{yqhuang2912@gmail.com}

\author{Zerui Shao}
\orcid{0000-0002-4936-031X}
\affiliation{%
  \institution{Sichuan University}
  \department{School of Cyber Science and Engineering}
  \city{Chengdu}
  \state{Sichuan}
  \country{China}
}
\email{shaozerui@stu.scu.edu.cn}

\author{Ziyuan Yang}
\orcid{0000-0002-0275-4098}
\affiliation{%
  \institution{Sichuan University}
  \department{College of Computer Science}
  \city{Chengdu}
  \state{Sichuan}
  \country{China}
}
\email{cziyuanyang@gmail.com}

\author{Zexin Lu}
\orcid{0000-0001-5307-145X}
\affiliation{%
  \institution{Sichuan University}
  \department{College of Computer Science}
  \city{Chengdu}
  \state{Sichuan}
  \country{China}
}
\email{zexinlu.scu@gmail.com}

\author{Yi Zhang}
\orcid{0000-0001-7201-2092}
\affiliation{%
  \institution{Sichuan University}
  \department{School of Cyber Science and Engineering}
  \city{Chengdu}
  \state{Sichuan}
  \country{China}
}
\email{yzhang@scu.edu.cn}


\begin{abstract}
  Mobile and Web-of-Things (WoT) devices at the network edge generate vast amounts of data for machine learning applications, yet privacy concerns hinder centralized model training. Federated Learning (FL) allows clients (devices) to collaboratively train a shared model coordinated by a central server without transfer private data, but inherent statistical heterogeneity among clients presents challenges, often leading to a dilemma between clients' needs for personalized local models and the server's goal of building a generalized global model. Existing FL methods typically prioritize either global generalization or local personalization, resulting in a trade-off between these two objectives and limiting the full potential of diverse client data. To address this challenge, we propose a novel framework that simultaneously enhances global generalization and local personalization by \textbf{R}ethinking \textbf{I}nformation \textbf{R}epresentation in the \textbf{Fed}erated learning process (\textbf{FedRIR}). Specifically, we introduce Masked Client-Specific Learning (MCSL), which isolates and extracts fine-grained client-specific features tailored to each client's unique data characteristics, thereby enhancing personalization. Concurrently, the Information Distillation Module (IDM) refines the global shared features by filtering out redundant client-specific information, resulting in a purer and more robust global representation that enhances generalization. By integrating the refined global features with the isolated client-specific features, we construct enriched representations that effectively capture both global patterns and local nuances, thereby improving the performance of downstream tasks on the client. Extensive experiments across diverse datasets demonstrate that FedRIR significantly outperforms state-of-the-art FL methods, achieving up to a 3.93\% improvement in accuracy while ensuring robustness and stability in heterogeneous environments. The code is available at \href{https://github.com/Deep-Imaging-Group/FedRIR}{https://github.com/Deep-Imaging-Group/FedRIR}.
\end{abstract}




\maketitle

\section{Introduction}
Mobile and Web-of-Things (WoT) devices at the network edge now generate massive amounts of data daily~\cite{wang2023flexifed,deng2023hsfl}, which recent advancements in deep learning have demonstrated to be crucial for achieving high performance in data-driven models~\cite{deng2009imagenet,devlin2018bert,kaplan2020scaling}. However, data at individual clients is often scarce and biased, making it challenging to build accurate and robust models~\cite{dhruva2020aggregating,chang2020synthetic}. Aggregating data from multiple clients into a centralized dataset might seem like a natural solution, but privacy and security concerns associated with sharing sensitive data across clients significantly hinder this approach~\cite{nguyen2021federated,guan2024federated}.

Federated Learning (FL) offers a promising solution to these challenges by enabling collaborative model training across multiple clients without the need to transfer local data~\cite{mcmahan2017communication,wang2024feddse}. In FL, each client retains its private data and trains a local model, sharing only model parameters with a central server. This approach addresses privacy concerns while leveraging collective data to improve overall model performance. However, client data is typically non-IID~\cite{li2020federated,zhang2023fedala,zhang2024few}, and the heterogeneity issue poses a significant challenge for traditional FL methods, such as FedAvg, as a single global model often fails to adequately capture the diverse data characteristics, leading to suboptimal performance across clients~\cite{huang2021personalized,t2020personalized}.

Personalized Federated Learning (pFL) was introduced to tackle data heterogeneity across clients. It aims to build models tailored to each client by leveraging client-specific data variations to improve overall performance~\cite{hanzely2020lower,t2020personalized}. In pFL, each client maintains a personalized component that adapts to its local data, while also contributing the parameters of a global component aimed at extracting refined global features. These global parameters are then aggregated on the server to improve generalization and address challenges associated with data scarcity. However, existing pFL methods often prioritize either global information representation, such as in FedRoD~\cite{chen2021bridging}, or personalized information representation, as demonstrated by FedPer~\cite{arivazhagan2019federated} and FedRep~\cite{pmlr-v139-collins21a}, during local training. This singular focus tends to overlook the complementary aspect, limiting the ability to effectively balance collaborative learning with personalization. Although methods like FedProto~\cite{tan2022fedproto} attempt to bridge this gap by employing prototypes to guide personalized feature extraction, their effectiveness heavily relies on the quality of global representations. Poor global features can misguide the personalization process, leading to suboptimal performance in both global and personalized objectives~\cite{zhang2023fedcp}. The recently proposed FedAKD~\cite{DING2024112317} seeks to optimize both global and personalized models through server-side and client-side knowledge distillation, but it relies on a reference dataset accessible to both the server and clients, which is often impractical in real-world scenarios. Therefore, it is crucial to develop methods that can simultaneously enhance global generalization and local personalization within federated learning.

To meet this need, we propose FedRIR, an innovative pFL framework that enhances both local personalization and global generalization by jointly optimizing two key components: Masked Client-Specific Learning (MCSL) and the Information Distillation Model (IDM). MCSL isolates and extracts client-specific features by applying masking techniques that emphasize the unique characteristics of each client's data. IDM refines the shared global features by filtering out redundant or overly client-specific information, producing a purer and more robust global representation that captures common patterns across all clients. By integrating these components, FedRIR achieves an effective balance between personalization and generalization. In summary, the main contributions of this paper are as follows:

\begin{itemize} 
    \item Propose a Masked Client-Specific Learning mechanism that effectively extracts client-specific heterogeneous information, enhancing model personalization.
    \item Develop an Information Distillation Module that refines global features, resulting in a purer global representation and optimizing the model’s capacity for improved generalization.
    \item Extensive experiments demonstrate that FedRIR not only achieves superior performance and communication efficiency but also maintains robustness and stability across various heterogeneous scenarios, outperforming state-of-the-art methods.
\end{itemize}

\section{Related Work}
\subsection{Personalized Federated Learning}
Personalized Federated Learning (pFL) has been proposed to address statistical heterogeneity and improve personalization in FL by training a unique, personalized model for each client rather than relying on a single global model~\cite{t2020personalized,yi2023fedgh}. 

Existing pFL approaches can be grouped into five main categories~\cite{zhang2023pfllib}: (1) Meta-learning-based pFL: Methods like Per-FedAvg~\cite{NEURIPS2020_24389bfe} fine-tune the global model using local data, resulting in a more personalized model for each user. Although this method enhances personalization, it may face challenges in maintaining consistency across clients due to local data variations. (2) Regularization-based pFL: Ditto~\cite{pmlr-v139-li21h} uses a proximal term to incorporate global information from the global model parameters during local training. This approach helps balance personalization and global consistency, but it may not fully capture the individual nuances of each client's data. (3) Personalized-aggregation-based pFL: Approaches like FedALA~\cite{zhang2023fedala} use an Adaptive Local Aggregation module to dynamically merge the global model with the local model for each client. Although this technique is effective, its success depends on how well the aggregation aligns with each client’s specific needs, which can be challenging in diverse settings. (4) Model-splitting-based pFL: Methods such as FedRep, FedPer~\cite{arivazhagan2019federated}, and FedRoD split the model into a global feature extractor and a client-specific head. These approaches focus on enhancing either global or personalized feature representation, potentially limiting overall model performance by not fully integrating both aspects. (5) Knowledge-distillation-based pFL: FedKD~\cite{wu2022communication} utilizes knowledge distillation to train a student model guided by a teacher model, sharing only the student model to significantly reduce communication costs. Although this method is efficient, it may not adequately balance global and personalized information representation.

Despite advancements, existing pFL methods often prioritize either global or personalized information representation during local training. This focus limits their ability to achieve a balance between collaborative learning and personalization, especially in diverse and heterogeneous data environments.

\subsection{Masked Representation Learning} 
Masked representation learning is a self-supervised technique that predicts masked components using contextual information, allowing models to learn rich and meaningful representations without labeled data. This approach has gained widespread adoption across various domains, especially in Natural Language Processing (NLP) and Computer Vision (CV)~\cite{devlin2018bert,he2022masked,fang2023eva}.

In NLP, masked language models have transformed how machines understand and generate human language. The groundbreaking work by Devlin \etal~\cite{devlin2018bert} introduced BERT, which uses masked language modeling to capture bidirectional context. By randomly masking a subset of input tokens and training the model to predict them, BERT effectively learns deep contextual representations that serve as the foundation for various downstream tasks, including question answering, sentiment analysis, and machine translation. Similarly, in CV, masked representation learning has been successfully adapted to image data. He \etal~\cite{he2022masked} proposed the Masked Autoencoder (MAE), which improves efficiency by sparsely applying the ViT~\cite{dosovitskiy2020image} encoder to the visible (unmasked) content only. MAE employs high masking ratios, typically masking 75\% of the input patches, thereby reducing computational overhead and promoting the learning of more robust and generalizable features from the visible data. Beyond NLP and CV, masked representation learning has been applied in other domains such as audio processing~\cite{niizumi2022masked}, where models learn audio representations by auto-encoding masked spectrogram patches, and graph neural networks~\cite{li2023s}, where masking nodes or edges helps facilitate the learning of structural and relational information.

MCSL extends masked representation learning to federated settings by focusing on extracting client-specific representations finely tuned to each client’s unique data characteristics. By utilizing a proper masking ratio, MCSL enhances client-specific feature extraction while mitigating the risk of overfitting to local data. As a crucial component in improving personalization within pFL, MCSL leverages the strengths of masked learning to develop more efficient and privacy-preserving models in federated environments.
 
\section{Methodology}
\subsection{Overview}
\begin{figure}[t]
    \centering
    \includegraphics[width=0.5\textwidth]{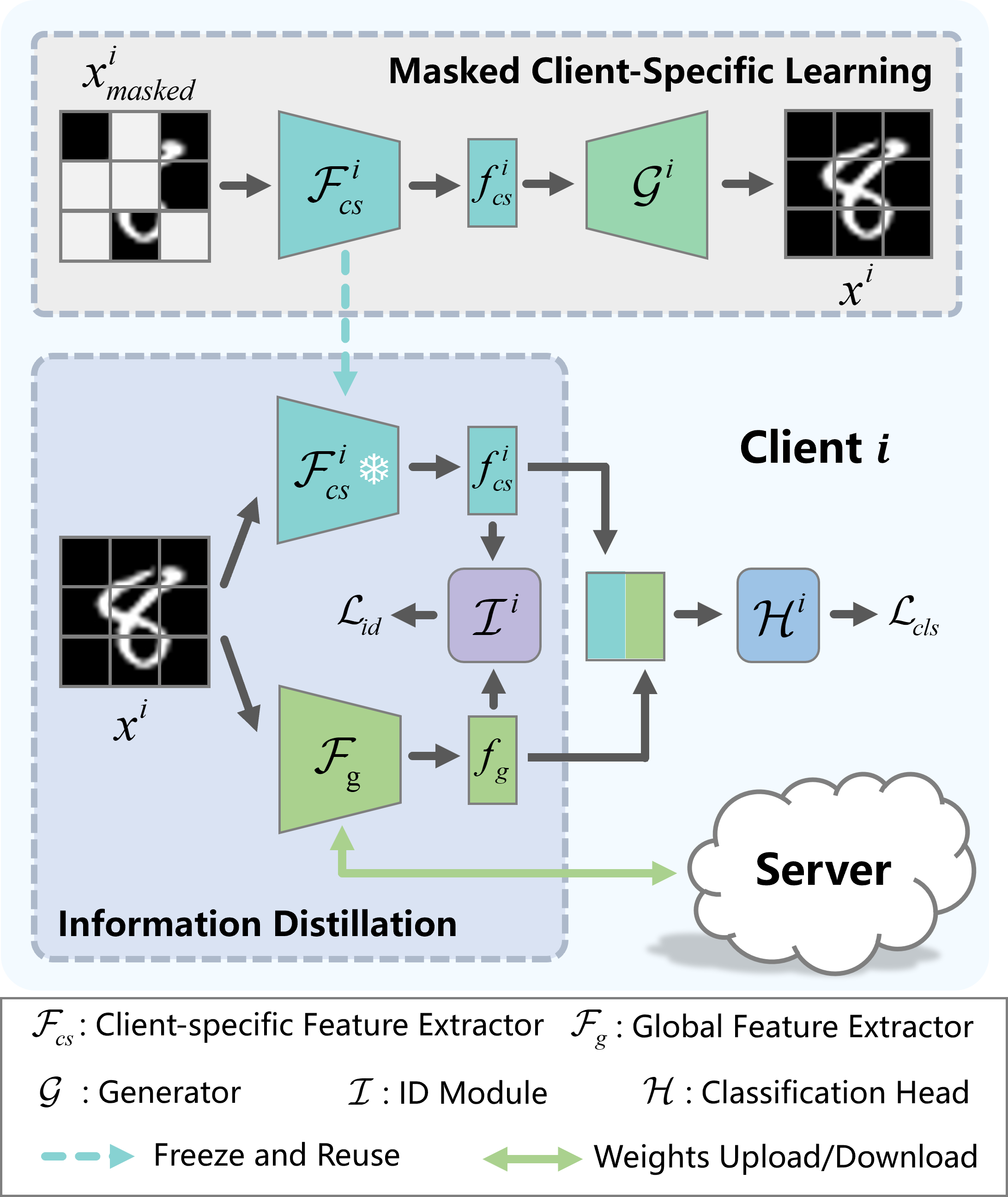}
    \caption{The workflow of the proposed FedRIR framework.}
    \Description{A diagram showing the workflow of the proposed FedRIR framework.}
    \label{workflow}
\end{figure}

In pFL, we consider $N$ clients, each with its own privacy-sensitive dataset $\mathcal{D}^i$ exhibiting statistical heterogeneity. The goal of the proposed FedRIR framework is to improve both the generalization and personalization of the overall model by refining the global and client-specific representations. This is achieved through iterative local client updates and server aggregation at each communication round. The FedRIR workflow is shown in Figure \ref{workflow}.

\subsubsection{Local Client Update}
As shown in Figure \ref{workflow}, the FedRIR training process consists of two key stages: Masked Client-Specific Learning (MCSL) and Information Distillation (ID).

In the MCSL stage, client-specific representations are learned. For each client $i$, the input data $x^{i}$ undergoes a masking operation, resulting in masked data $x^{i}_{masked}$. The masked data is then processed by a client-specific feature extractor $\mathcal{F}^{i}_{cs}$ to obtain client-specific features $f^{i}_{cs}$. These features are subsequently used by a generator $\mathcal{G}^{i}$ to reconstruct the original input $x^{i}$.

The ID stage focuses on refining global representations by filtering out redundant information. At this stage, the client-specific feature extractor $\mathcal{F}^{i}_{cs}$ is frozen to preserve the learned parameters. The original input $x^{i}$ is processed by the frozen $\mathcal{F}^{i}_{cs}$ to yield $f^{i}_{cs}$ and by the global feature extractor $\mathcal{F}_{g}$ to produce global features $f_{g}$. An Information Distillation Module (IDM) is then applied to eliminate redundant client-specific details from the global features by minimizing the mutual information between the global features and the client-specific features.

Finally, a classification head $\mathcal{H}^{i}$ maps the concatenated features $f_{g}$ and $f^{i}_{cs}$, forming the personalized features $f^{i}_{p} = \text{concat}(f_{g}, f^{i}_{cs})$, to the classification label space on the client. This mapping, represented as $\hat{y}^{i} = \mathcal{H}^{i}(f^{i}_{p})$, integrates the refined global features with the client-specific features to construct enriched representations that capture both global patterns and local nuances, enabling accurate predictions.

\subsubsection{Server Aggregation}
After each round of local updates, each client $i$ sends the learned parameters of its global feature extractor ${\mathcal{F}}_{g}$ to the server for aggregation. Following a similar approach to FedAvg, the aggregation process is defined as:

\begin{equation}
\bar{\mathcal{W}}_{\mathcal{F}_{g}} = \sum_{i=1}^{N}\frac{|\mathcal{D}^i|}{\sum_{i=1}^{N}|\mathcal{D}^i|} \mathcal{W}^{i}_{\mathcal{F}_{g}}
\label{eq_agg}
\end{equation}

Here, $\bar{\mathcal{W}}_{\mathcal{F}_{g}}$ represents the globally aggregated parameters of all participating global feature extractors, and $|\mathcal{D}^i|$ denotes the number of data samples on the $i$-th client. After $\bar{\mathcal{W}}_{\mathcal{F}_{g}}$ is computed, it is broadcast to all clients. Each client then updates its local global feature extractor $\mathcal{F}_{g}$ by replacing its parameters with the globally aggregated parameters $\bar{\mathcal{W}}_{\mathcal{F}_{g}}$.

\subsection{Masked Client-Specific Learning}

In FL, an effective client-specific feature extractor should accurately capture the unique characteristics of each client's data without simply overfitting to the training samples. To achieve this, we introduce Masked Client-Specific Learning (MCSL). MCSL encourages the client-specific feature extractor to focus on truly client-specific aspects of the data by masking parts of the input, which helps isolate and highlight the distinctive features of each client's dataset. This process ensures that the client-specific feature extractor is finely tuned to the client's unique data distribution, while also avoiding overfitting.

Capturing the unique aspects of local data $\mathcal{D}^{i}=\{(x_{k}^{i}, y_{k}^{i})\}_{k=1}^{|\mathcal{D}^{i}|}$ on client $i$ can be theoretically framed as maximizing the mutual information between the input data $x^{i}$ and the client-specific representation $f_{cs}^{i}$~\cite{vincent2010stacked}. Maximizing this mutual information encourages the feature extractor to retain as much relevant information about the input on client $i$ as possible, thereby enhancing personalization. This can be expressed as:
\begin{equation}
\begin{aligned}
    \mathop{\arg \max} I(x^{i}; f_{cs}^{i}) = \mathop{\arg \max} \mathbb{E}_{p(x^{i}, f_{cs}^{i})}\left[\log \frac{p(x^{i}|f_{cs}^{i})}{p(x^{i})}\right]
\end{aligned}
\end{equation}
Since directly modeling the conditional distribution $p(x^{i}|f_{cs}^{i})$ is intractable, we adopt an approach based on the Barber-Agakov (BA) bound~\cite{barber2004algorithm}, introducing an auxiliary distribution $q(x^{i}|f_{cs}^{i})$ to derive a tractable lower bound for the mutual information $I(x^{i}; f_{cs}^{i})$:
\begin{equation}
    I_{\text{BA}} := H(x^i) + \mathbb{E}_{p(x^{i}, f_{cs}^{i})}\left[\log q(x^{i}|f_{cs}^{i})\right] \leq I(x^{i}; f_{cs}^{i})
    \label{eq_I_BA}
\end{equation}
Here, $H(x^i)=\mathbb{E}_{p(x^i)}[-\log p(x^i)]$ is the entropy of the input data $x^i$, which depends solely on the data collection process and is independent of the feature extraction process. Consequently, we focus on maximizing the second term of $I_{\text{BA}}$, which represents the process of reconstructing the original input $x^i$ from the client-specific features $f_{cs}^{i}$. To achieve this, we employ a variational autoencoder (VAE)~\cite{kingma2013auto} framework, constructing an encoder $\mathcal{F}_{cs}^i$ to extract the client-specific features $f_{cs}^{i}$ and a generator $\mathcal{G}^{i}$ to reconstruct the input data $x^i$. The loss function to minimize the reconstruction error and thereby maximize $I(x^{i}; f_{cs}^{i})$ is:
\begin{equation}
    \mathcal{L}_{recon}=\left\|\mathcal{G}^{i}(f_{cs}^{i})-x^i\right\|_{2}  
    \label{l_recon1}
\end{equation}

However, previous researches~\cite{vincent2010stacked,he2022masked} indicate that direct reconstruction from the original input $x^i$ can lead to trivial solutions, such as simply copying the input, which fails to capture meaningful representations. This issue is especially problematic in federated learning, where each client's data is typically heterogeneous and limited in size. To address this, we introduce random masking of portions of the input data, preventing the client-specific feature extractor $\mathcal{F}_{cs}^{i}$ from overfitting. This effectively transforms the VAE into a denoising autoencoder (DAE)~\cite{vincent2010stacked}, with the modified reconstruction loss:
\begin{equation}
    \mathcal{L}_{recon}=\left\|\mathcal{G}^{i}\left(\mathcal{F}_{cs}^{i}(x_{masked}^{i})\right)-x^i\right\|_{2}  
    \label{l_recon2}
\end{equation}

\subsection{Information Distillation}

Once the client-specific features $f_{cs}^{i}$ have been extracted, a global feature extractor $\mathcal{F}_{g}$ is used to derive global features $f_{g}$ from the input sample $x^i$. To ensure that the global feature extractor effectively captures pure and generalized features, we introduce an Information Distillation (ID) strategy. This strategy distills out redundant client-specific information from the global features, leading to two key benefits: (1) reducing the influence of non-common noise (unexpected client-specific noise) when the parameters of the global feature extractor are aggregated on the server, and (2) minimizing redundancy between the client-specific and global features, thereby improving the performance of subsequent downstream tasks when these features are concatenated. By reducing the mutual information $I(f_{g}, f_{cs}^i)$, we ensure that the global and client-specific features capture distinct aspects of the data, minimizing redundancy. This enhances the informativeness of both representations, leading to improved model performance. The ID process is formally defined as:
\begin{equation}
\begin{aligned}
    \mathop{\arg \min} I(f_{cs}^i, f_{g}) = \mathop{\arg \min} \mathbb{E}_{p(f_{cs}^i, f_{g})}\left[\log \frac{p(f_{g}|f_{cs}^{i})}{p(f_{g})}\right]
\end{aligned}
\label{eq_I_cp_cs}
\end{equation}

However, directly solving Eq. (\ref{eq_I_cp_cs}) is challenging due to the intractability of the conditional distribution $p(f_{g}|f_{cs}^{i})$. To address this, we employ a practical approach using a variational approximation $q_{\theta}(f_{g}|f_{cs}^{i})$, as proposed in~\cite{belghazi2018mutual,cheng2020club}, to estimate a reliable upper bound of the mutual information. Specifically, we utilize the Variational Constrastive Log-ratio Upper Bound (vCLUB)~\cite{cheng2020club} to derive this bound:
\begin{equation}
\begin{aligned}
    I_{\text{vCLUB}} &:= \mathbb{E}_{p(f_{cs}^i,f_{g})}\left[\log q_{\theta}(f_{g}|f_{cs}^{i})\right] \\
    &- \mathbb{E}_{p(f_{cs}^i)}\mathbb{E}_{p(f_{g})}\left[\log q_{\theta}(f_{g}|f_{cs}^{i})\right] \\
    &\geq I(f_{cs}^i, f_{g})
\end{aligned}
\end{equation}

By minimizing this upper bound, the redundancy between $f_{g}$ and $f_{cs}^i$ is significantly reduced, resulting in more distinct and informative global and client-specific features. In practice, $q_{\theta}(f_{g}|f_{cs}^{i})$ is typically implemented with neural networks, and we introduce an Information Distillation Module $\mathcal{I}^i$ to implement and train $q_{\theta}(f_{g}|f_{cs}^{i})$ with the following loss function:
\begin{equation}
    \mathcal{L}_{id} = I_{\text{vCLUB}}(f_{cs}^i, f_{g})
\end{equation}

Then, the extracted client-specific features $f_{cs}^i$ and the refined global features $f_{g}$ are concatenated into personalized features $f_{p}^i = [f_{cs}^i, f_{g}]$, which is fed into a classification head $\mathcal{H}^i$ to compute the personalized classification output. This classification process is trained with the following loss functions:
\begin{equation}
    \mathcal{L}_{cls} = \text{CrossEntropy}\left(\mathcal{H}^i(f_{p}^i), y^i\right)
\end{equation}
Thus, the overall loss function at client $i$ is:
\begin{equation}
    \mathcal{L} = \mathcal{L}_{id} + \mathcal{L}_{cls}
    \label{l_total}
\end{equation} 
In our implementation, we set the balance hyperparameters between the different loss components to 1, and the full learning procedure is detailed in Algorithm \ref{algo_1}.

\begin{algorithm}
    \caption{The learning procedure of FedRIR}
    \begin{algorithmic}[1]
        \Require $N$ clients $\{C^i\}_{i=1}^{N}$ with their local data $\{\mathcal{D}^i\}_{i=1}^{N}$; initial weights of $\mathcal{F}_{cs}^{i},\mathcal{G}^{i},\mathcal{F}_{g},\mathcal{H}^{i},\mathcal{I}^{i}$; masked ratio $r$; client joining ratio $\rho$; total communication rounds $T$.
        \Ensure Trained model parameters of $\mathcal{F}_{g},\{\mathcal{F}_{cs}^{i},\mathcal{H}^{i}\}_{i=1}^{N}$.
        \State Each client $i$ initializes its local models $\mathcal{F}_{cs}^{i},\mathcal{G}^{i},\mathcal{F}_{g},\mathcal{H}^{i},\mathcal{I}^{i}$.
        \For{iteration $t=0,\cdots,T$}
            \State Server samples a clients subset $\{C^i\}_{i=1}^{\rho \times N}$.
            \State Server broadcasts $\bar{\mathcal{W}}_{\mathcal{F}_{g}}$ to all selected clients.
            \For{Client $C^{i} \in \{C^i\}_{i=1}^{\rho \times N}$ in parallel}
                \Statex \Comment{Masked Client-Specific Learning}
                \State Mask the local data $x^i$ with ratio $r$ to obtain $x_{masked}^{i}$.
                
                \State Optimize $\mathcal{F}_{cs}^{i}$ and $\mathcal{G}^{i}$ on $x_{masked}^{i}$ by minimizing $\mathcal{L}_{recon}$.
                \Statex \Comment{Information Distillation}
                \State Freeze $\mathcal{F}_{cs}^{i}$.
                \State Compute $f_{cs}^{i} = \mathcal{F}_{cs}^{i}(x^i)$ and $f_{g} = \mathcal{F}_{g}(x^i)$.
                \State Update $\mathcal{F}_{g}$, $\mathcal{H}^{i}$ and $\mathcal{I}^{i}$ by minimizing $\mathcal{L}=\mathcal{L}_{id}+\mathcal{L}_{cls}$.
            \EndFor
            \Statex \Comment{Server Aggregation}
            \State Server collects $\mathcal{W}^{i}_{\mathcal{F}_{g}}$ from all selected clients.
            \State Server computes $\bar{\mathcal{W}}_{\mathcal{F}_{g}}$ using Eq. (\ref{eq_agg}).
        \EndFor
        \State \Return Trained model parameters of $\mathcal{F}_{g},\{\mathcal{F}_{cs}^{i},\mathcal{H}^{i}\}_{i=1}^{N}$.
    \end{algorithmic}
    \label{algo_1}
\end{algorithm}

\section{Experiments}
\subsection{Statistically Heterogeneous Settings}
\begin{figure}[ht]
    \centering
    \includegraphics[width=0.5\textwidth]{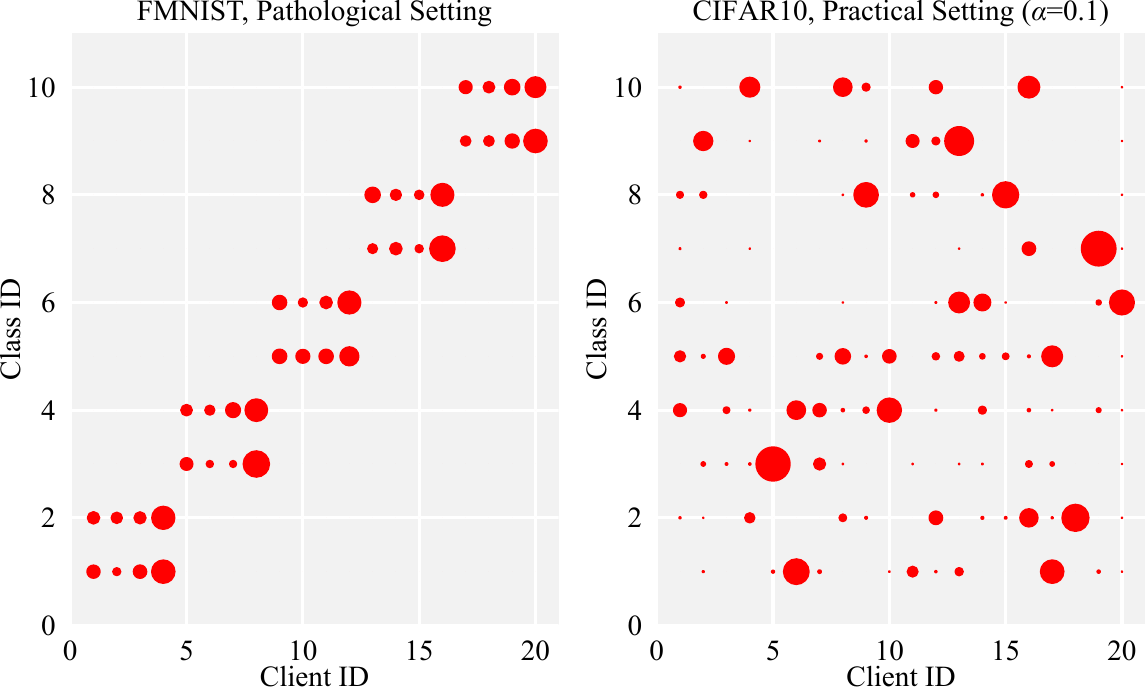}
    \caption{The data distribution on FMNIST and Cifar10 datasets in pathological and practical settings with 20 clients. The size of a circle indicates the number of local data.}
    \Description{}
    \label{fig:dataset}
\end{figure}

We evaluate model performance under various heterogeneous data scenarios using three settings: pathological, practical, and real-world. These settings simulate different degrees of statistical heterogeneity commonly observed in federated learning environments. Figure \ref{fig:dataset} illustrates the data distribution on FMNIST~\cite{xiao2017fashion} and Cifar10~\cite{krizhevsky2009learning} datasets in the pathological and practical settings with 20 clients.

\subsubsection{Pathological Setting.} This setting simulates extreme heterogeneity by assigning each client a subset of classes, resulting in non-overlapping labels among clients. For Cifar10 and Cifar100~\cite{krizhevsky2009learning}, each client receives data from only 2 out of 10 and 10 out of 100 classes, respectively, following the approach of FedCP~\cite{zhang2023fedcp}. This setup emphasizes the model's ability to learn in highly fragmented environments.

\subsubsection{Practical Setting.} This setting evaluates performance in scenarios that closely resemble real-world data distribution using the Dirichlet distribution, as implemented in previous works~\cite{li2021model,zhang2023fedcp}. This setting is applied to the Cifar10 and MNIST~\cite{lecun1998gradient} datasets, and we sample $q_{c,i}\sim Dir(\alpha)$ to determine the proportion of samples from class $c$ allocated to client $i$. For our experiments, without explicitly stated, $\alpha$ is set to $0.1$. 

\subsubsection{Real-world Setting.} The real-world setting involves heterogeneous datasets with significant differences in data distributions across clients due to varying acquisition conditions. We utilize two datasets: (1) OfficeCaltech10~\cite{ghifary2016scatter}, which includes four distinct data sources, three from Office-31 and one from Caltech-256, each acquired using different camera devices or in different environments with various backgrounds. (2) DomainNet~\cite{peng2019moment}, which comprises images from six different sources. Each client in our experiments is assigned data from only one of these sources, resulting in non-IID data across clients.

\subsection{Implementation Details}

Following the setup in FedRoD, unless otherwise specified, our experiments involve 20 clients with a participation ratio of $\rho = 1$, and each client's data is divided into a training set (75\%) and a test set (25\%). Based on our hyper-parameter analysis, we set the masking ratio $r = 0.6$. Consistent with previous works~\cite{luo2021no,zhang2023gpfl}, we use a simple CNN for the feature extractors $\mathcal{F}$, composed of two convolutional layers, each followed by Batch Normalization, ReLU activation, and Max Pooling layers. The classification head $\mathcal{H}$ includes a fully connected layer, while the information distillation module $\mathcal{I}$ consists of four fully connected layers, each followed by a ReLU activation function. The generator $\mathcal{G}$ uses two deconvolution layers. Additionally, each task is trained for 1000 communication rounds using the Adam optimizer with a learning rate of $5 \times 10^{-4}$, a batch size of 100, and one epoch per local training round. Additionally, all experiments are conducted on a machine equipped with an Intel Xeon Gold 5220 CPU @ 2.20GHz and an NVIDIA GeForce RTX 3090 GPU. Additionally, each experiment is repeated three times to ensure statistical significance, and we report the mean and standard deviation of the results.

\begin{table*}[ht]
\setlength{\tabcolsep}{5pt}  
\centering
\caption{The average test accuracy (\%) across different federated learning methods in various heterogeneous scenarios, along with the number of communication parameters used by each method.}
\begin{tabular}{l|cc|cc|cc|c}
\hline \toprule
\textbf{Scenarios}      & \multicolumn{2}{c|}{\textbf{Pathological}} & \multicolumn{2}{c|}{\textbf{Practical ($\alpha=0.1$)}} & \multicolumn{2}{c|}{\textbf{Real-world}} & \textbf{$\#$ of Comm. Params.} \\ \midrule
\textbf{Datasets}       & Cifar100       & Cifar10   & MNIST          & Cifar10        & OfficeCaltech10   & DomainNet         & -   \\ \midrule
Local Training    & 69.41$\pm$0.22 & 86.36$\pm$0.04 & 99.16$\pm$0.02 & 89.27$\pm$0.25 & 68.66$\pm$0.34    & 57.99$\pm$0.03    & -           \\ \midrule
FedAvg                  & 26.50$\pm$0.55 & 86.01$\pm$0.15 & 97.49$\pm$0.32 & 53.41$\pm$0.34 & 68.14$\pm$0.45    & 54.92$\pm$0.33    & 5.597M       \\
Per-FedAvg              & 56.61$\pm$0.19 & 86.23$\pm$0.16 & 98.30$\pm$0.24 & 85.93$\pm$0.12 & 66.61$\pm$0.56    & 50.46$\pm$0.64    & 5.597M       \\
FedProto                & 68.08$\pm$0.27 & 84.08$\pm$0.04 & 99.18$\pm$0.02 & 86.94$\pm$0.16 & 66.81$\pm$0.56    & 55.94$\pm$0.18    & \textbf{5.120K}       \\
Ditto                   & 69.44$\pm$0.21 & 86.14$\pm$0.05 & 99.15$\pm$0.03 & 89.61$\pm$0.11 & 68.98$\pm$0.13    & 55.02$\pm$0.14    & 5.597M       \\
FedRep                  & 67.97$\pm$0.32 & 85.86$\pm$0.07 & 99.34$\pm$0.03 & \underline{90.14$\pm$0.03} & 69.50$\pm$0.59    & 58.38$\pm$0.03    & 5.592M       \\
FedRoD                  & 64.04$\pm$0.65 & 86.43$\pm$0.20 & \underline{99.49$\pm$0.02} & 89.64$\pm$0.03 & \underline{69.66$\pm$0.63}    & 57.49$\pm$0.39   & 5.597M \\
FedBABU                 & 60.34$\pm$0.14 & 84.26$\pm$0.25 & 98.92$\pm$0.00 & 88.05$\pm$0.14 & 68.24$\pm$0.61    & 56.16$\pm$0.38    & 5.592M         \\
FedALA                  & 63.31$\pm$0.48 & \underline{86.65$\pm$0.01} & 99.32$\pm$0.01 & 89.42$\pm$0.11 & 67.77$\pm$0.65    & 55.92$\pm$0.39    & 5.597M         \\
FedKD                   & \underline{69.61$\pm$0.17} & 86.12$\pm$0.01 & 98.81$\pm$0.03 & 89.77$\pm$0.11 & 68.19$\pm$1.05    & 57.20$\pm$0.03    & \underline{5.590M} \\
FedCP                   & 65.15$\pm$0.31 & 85.85$\pm$0.09 & 99.44$\pm$0.05 & 89.86$\pm$0.04 & 67.82$\pm$0.41    & \underline{59.02$\pm$0.02}    & 6.124M \\ \midrule
\textbf{FedRIR}  & \textbf{72.07$\pm$0.12} & \textbf{90.58$\pm$0.07} & \textbf{99.69$\pm$0.01} & \textbf{91.64$\pm$0.03} & \textbf{73.23$\pm$0.59} & \textbf{61.92$\pm$0.12} & 5.592M \\  \midrule
$\Delta$ \text{SOTA}    & $\uparrow$ 2.46 & $\uparrow$ 3.93 & $\uparrow$ 0.20 & $\uparrow$ 1.50 & $\uparrow$ 3.57 & $\uparrow$ 2.90    &  \\ \bottomrule
\end{tabular}
\label{table_effectiveness}
\end{table*}

\subsection{Performance Comparison}

We evaluate the performance of FedRIR in terms of effectiveness, scalability, stability, and hyper-parameter impact, comparing it with eleven state-of-the-art federated learning methods. These include Local Training (each client trains independently without server communication), FedAvg~\cite{mcmahan2017communication}, per-FedAvg~\cite{NEURIPS2020_24389bfe}, FedProto~\cite{tan2022fedproto}, Ditto~\cite{pmlr-v139-li21h}, FedRep~\cite{pmlr-v139-collins21a}, FedRoD~\cite{chen2021bridging}, FedBABU~\cite{oh2022fedbabu}, FedALA~\cite{zhang2023fedala}, FedKD~\cite{wu2022communication}, and FedCP~\cite{zhang2023fedcp}. The results across diverse heterogeneous scenarios are shown in Table~\ref{table_effectiveness}.

\subsection{Effectiveness}
\begin{figure}[ht]
    \centering
    \includegraphics[width=0.5\textwidth]{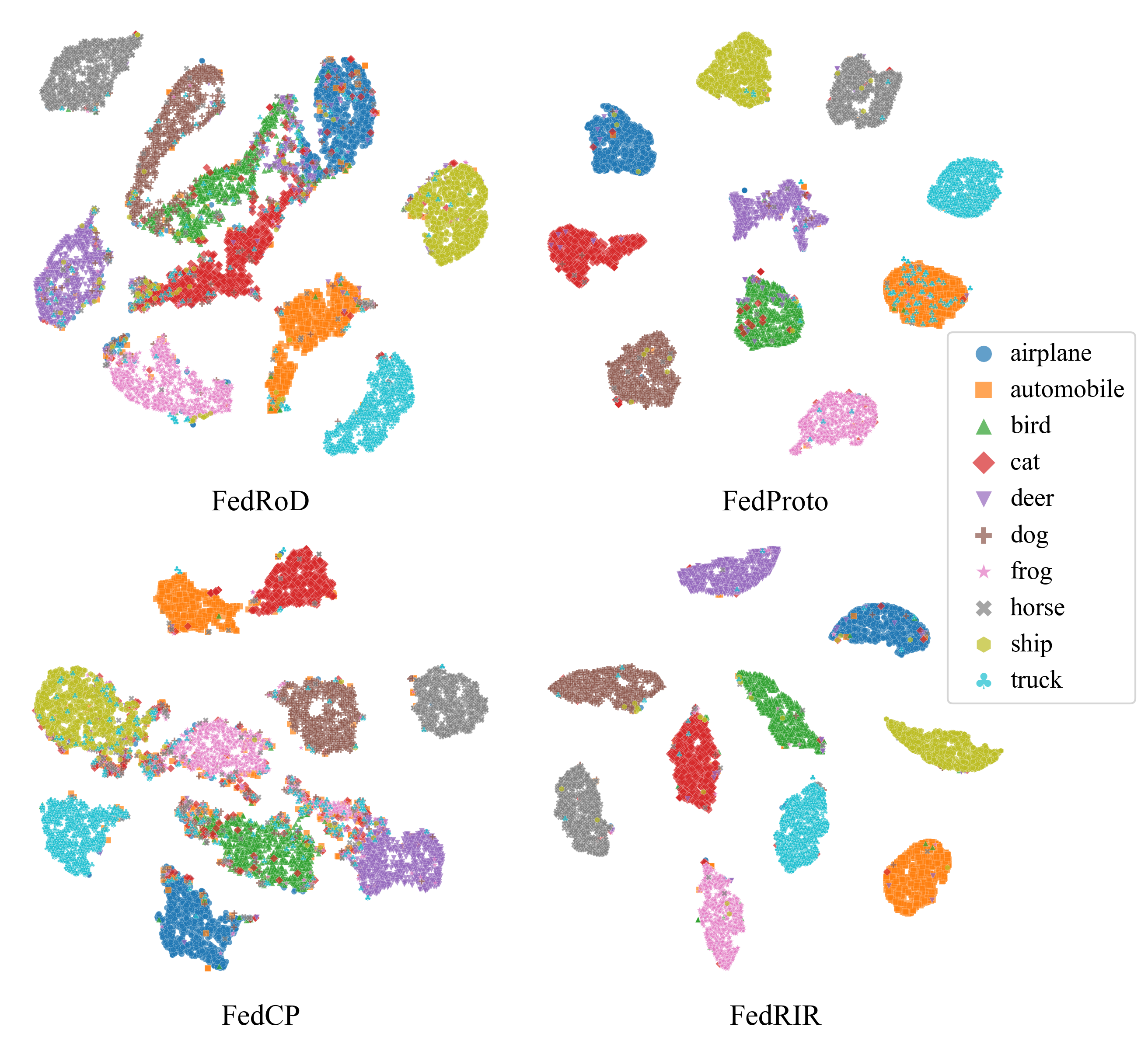}
    \caption{t-SNE visualization of global features extracted by FedRoD, FedProto, FedCP and FedRIR on Cifar10 dataset, with color represents a different class.}
    \Description{}
    \label{fig:global_features}
\end{figure}

\subsubsection{Global Generization Effectiveness} 
 As demonstrated in Table~\ref{table_effectiveness}, FedRIR consistently outperforms Local Training, effectively capturing and utilizing shared knowledge among clients to enhance performance beyond what isolated local models can achieve. For instance, on the real-world datasets OfficeCaltech10 and DomainNet, FedRIR surpasses Local Training by 4.57\% and 3.93\%, respectively, highlighting its ability to leverage federated learning to improve generalization across diverse data distributions. When compared to exact global methods like FedAvg, FedRIR shows substantial advantages, particularly in non-IID environments where FedAvg struggles with conflicting updates from heterogeneous client data, resulting in poor generalization. For example, on Cifar100, FedAvg achieves only 26.50\%, significantly lower than FedRIR’s performance.

Figure~\ref{fig:global_features} provides further insight into the superior global feature separation achieved by FedRIR. When compared to pFL methods that emphasize global feature enhancement, such as FedRoD and FedProto, as well as methods that explore both global and personalized representations, such as FedCP, FedRIR demonstrates clearer and more distinct clustering across classes. The minimal overlap between different categories in FedRIR's global features reflects a well-separated and interpretable global representation. In contrast, FedRoD and FedCP show significant class overlap, indicating suboptimal global feature extraction. Although FedProto achieves clearer boundaries between some classes, misclassifications are evident, particularly where certain truck samples (light blue club suit symbols $\clubsuit$) are incorrectly clustered with automobiles (orange squares $\Box$). FedRIR’s use of information distillation strategy plays a critical role in addressing these issues by filtering out redundant client-specific information to refine the shared global model, ultimately resulting in superior generalization across clients.

\subsubsection{Local Personalization Effectiveness}
FedRIR stands out for its strong personalization capabilities, surpassing methods like Per-FedAvg, which struggles to balance local personalization and global accuracy. This issue is particularly evident in non-IID settings, where Per-FedAvg achieves only 56.61\% accuracy on Cifar100. In contrast, FedRIR combines individual client characteristics with refined global knowledge, enabling superior model adaptation and personalization across diverse data. Methods like Ditto fall short in personalization compared to FedRIR, as they struggle to capture the nuances of client-specific data. This limitation is evident in Ditto's lower accuracy on real-world datasets like DomainNet (55.02\%), whereas FedRIR achieves 61.92\%.

Similarly, approaches such as FedRep and FedPer, despite focusing on personalization, fail to fully leverage the global aggregated information from the federated learning paradigm to enhance personalization. This often results in neglecting essential global insights or facing conflicts between global and personalized objectives, leading to lower performance (e.g., FedRep: 58.38\% on DomainNet). Furthermore, FedALA’s adaptive local aggregation aims to merge global and local models, but its performance (67.77\% on OfficeCaltech10) is limited by the alignment between the aggregation and the specific needs of each client. FedKD uses knowledge distillation to reduce communication costs but compromises accuracy on diverse datasets, achieving only 57.20\% on DomainNet compared to FedRIR's 61.92\%. Similarly, FedCP separates global and personalized information without a robust refinement mechanism, leading to lower accuracy on Cifar100 (65.15\%) compared to FedRIR’s 72.07\%. These comparisons highlight FedRIR's ability to optimize global and personalized representations concurrently, consistently delivering superior personalization across diverse datasets.

\subsubsection{Communication Overhead Effectiveness}
In addition to its superior performance, FedRIR maintains efficient communication overhead, which is critical in federated learning environments. Despite incorporating more advanced personalization mechanisms, FedRIR’s communication costs are comparable to simpler methods like FedAvg, FedRep, and Ditto. By focusing on refining only the most essential features through MCSL and ID, FedRIR minimizes the amount of redundant or unnecessary data that needs to be transmitted between clients and the server. This makes it practical and scalable for real-world deployments, where communication efficiency is a crucial factor. For example, methods like FedKD, which uses knowledge distillation to reduce communication costs, may sacrifice model accuracy, especially in heterogeneous settings, whereas FedRIR achieves both high accuracy and comparable communication overhead, ensuring that its strong performance can be maintained in large-scale federated learning systems without imposing excessive communication burdens.

Overall, FedRIR's ability to simultaneously enhance generalization and personalization, while maintaining efficient communication, positions it as a robust and scalable solution for federated learning in diverse and challenging environments.

\begin{table}[ht]
\setlength{\tabcolsep}{2pt}  
\centering
\caption{The averaged test accuracy (\%) of Cifar100 classification for client amounts scalability.}
\begin{tabular}{l|cccc}
\hline \toprule
              & $N=10$          & $N=20$           & $N=50$           & $N=100$            \\  \midrule
FedAvg        & 30.26$\pm$0.02  & 30.10$\pm$0.12    & 21.17$\pm$0.05   & 14.14$\pm$0.06   \\
Per-FedAvg    & 45.54$\pm$0.03  & 43.96$\pm$0.04   & 37.43$\pm$0.08   & 33.58$\pm$0.09   \\
FedProto      & 48.09$\pm$0.02  & 46.15$\pm$0.06   & 44.80$\pm$0.06   & 41.68$\pm$0.08   \\
Ditto         & 46.33$\pm$0.04  & 45.24$\pm$0.05   & 43.28$\pm$0.10   & 41.09$\pm$0.10   \\
FedRep        & 49.60$\pm$0.03  & 47.87$\pm$0.09   & 45.06$\pm$0.08   & 40.88$\pm$0.09   \\
FedRoD        & 49.18$\pm$0.03  & 50.05$\pm$0.26   & 46.81$\pm$0.08   & 42.31$\pm$0.09  \\
FedBABU       & 49.42$\pm$0.02  & 49.63$\pm$0.29   & 43.36$\pm$0.07   & 37.20$\pm$0.06  \\
FedALA        & \underline{52.85$\pm$0.03}  & 48.63$\pm$0.04   & 45.09$\pm$0.08   & 40.66$\pm$0.09   \\
FedKD         & 51.68$\pm$0.04  & \underline{50.31$\pm$0.05}   & \underline{47.30$\pm$0.08}   & \underline{44.29$\pm$0.10}   \\
FedCP         & 50.21$\pm$0.01  & 48.38$\pm$0.47   & 45.89$\pm$0.04   & 41.85$\pm$0.07   \\ \midrule
\textbf{FedRIR} & \textbf{56.86$\pm$0.04}  & \textbf{54.75$\pm$0.05} & \textbf{53.63$\pm$0.03} & \textbf{50.02$\pm$0.04} \\  \midrule
$\Delta$ \text{SOTA} & $\uparrow$ 4.01 & $\uparrow$ 4.44 & $\uparrow$ 6.33 & $\uparrow$ 5.73  \\ \bottomrule
\end{tabular}
\label{table_scalability}
\end{table}
\subsection{Scalability}
To assess the scalability of FedRIR, we adopted the methodology from FedCP, partitioning the Cifar100 dataset into 10, 20, 50, and 100 sub-datasets to simulate 10, 20, 50, and 100 clients, respectively. As shown in Table~\ref{table_scalability}, FedRIR consistently outperforms state-of-the-art methods across various client numbers, maintaining strong performance even as the number of clients increases and data becomes increasingly fragmented. FedRIR’s accuracy remains robust, decreasing moderately from 56.86\% with 10 clients to 50.02\% with 100 clients, a decline of only 6.84 percentage points. In comparison, other methods suffer more pronounced performance drops.

As the number of clients grows, the performance gap between FedRIR and competing methods widens, highlighting FedRIR’s superior scalability. For example, with 10 clients, FedRIR outperforms the second-best method, FedALA, by 4.01 percentage points. However, this margin expands to 5.73 percentage points with 100 clients, demonstrating FedRIR's resilience against increasing client heterogeneity and data fragmentation.

This trend underscores the effectiveness of FedRIR’s design, particularly the combination of Masked Client-Specific Learning (MCSL) and Information Distillation (ID), which enable it to handle the complexities associated with scaling to larger numbers of clients more effectively than other approaches. The results affirm that FedRIR not only scales well but also becomes increasingly advantageous in large-scale federated learning environments, maintaining stability and accuracy where other methods falter.

\begin{table}[ht]
\centering
\caption{The averaged test accuracy (\%) of FMNIST classification with different participation ratios.}
\begin{tabular}{l|ccc}
\hline \toprule
\textbf{Algorithm} & \textbf{$\rho=1.0$} & \textbf{$\rho \in [0.5,1]$} & \textbf{$\rho \in [0.1,1]$} \\ \midrule
Per-FedAvg & 93.19$\pm$0.12 & 92.53$\pm$0.54 & 92.46$\pm$0.47 \\
FedProto & 92.19$\pm$0.04 & 91.77$\pm$0.08 & 90.36$\pm$0.57 \\
Ditto & \underline{95.98$\pm$0.01} & 95.36$\pm$0.43 & 95.02$\pm$1.01 \\
FedRep & 95.47$\pm$0.03 & 95.04$\pm$0.04 & 94.65$\pm$0.03 \\
FedROD & 95.57$\pm$0.02 & 95.22$\pm$0.32 & 94.48$\pm$0.93 \\
FedBABU & 95.21$\pm$0.13 & 95.21$\pm$0.13 & \underline{95.28$\pm$0.15} \\
FedALA & 95.56$\pm$0.00 & \underline{95.39$\pm$0.12} & 95.15$\pm$0.11 \\
FedKD & 95.17$\pm$0.03 & 94.92$\pm$0.05 & 94.21$\pm$0.08 \\
FedCP & 95.19$\pm$0.09 & 95.03$\pm$0.12 & 94.97$\pm$0.08 \\ \midrule
FedRIR & \textbf{97.51$\pm$0.00} & \textbf{97.48$\pm$0.04} & \textbf{97.43$\pm$0.02} \\ \midrule
$\Delta$ \text{SOTA} & $\uparrow$ 1.53 & $\uparrow$ 2.09 & $\uparrow$ 2.15   \\ \bottomrule
\end{tabular}
\label{table_stability}

\end{table}

\subsection{Stability}
We also evaluate the stability of our FedRIR in client dropout scenarios, which are commonly assessed in many pFL methods~\cite{zhang2023fedcp,oh2022fedbabu,zhang2023gpfl}. This situation is prevalent in real-world settings as some clients may accidentally drop out due to battery depletion or unstable network connections, resulting in not all clients participating in every communication round. Following the strategy used in previous work~\cite{zhang2023fedcp}, we vary the client joining ratio $\rho$ during each communication round on the FMNIST dataset in the training phase. Specifically, in our experiment, the client joining ratio $\rho$ is randomly sampled within a given range, such as $[0.5,1.0]$ and $[0.1,1.0]$, for each training epoch. As shown in Table \ref{table_stability}, although all methods exhibit performance drops as $\rho$ becomes more variable, FedRIR consistently maintains superior performance among the pFL methods, with the smallest performance decline. Additionally, the standard deviations of the results obtained by per-FedAvg, FedProto, Ditto, and FedRoD increase under unstable conditions, for example, when $\rho \in [0.1,1.0]$.

\subsection{Effect of Hyper-parameters}
\subsubsection{Effect of Mask Ratio $r$}
We conducted a series of experiments focusing on the Masked Client-Specific Learning (MCSL) component by applying various mask ratios. The results demonstrate three key observations: first, as the mask ratio increases, the test accuracy initially improves for all datasets, but after reaching an optimal point, the accuracy either stabilizes or slightly declines for some datasets. Second, using masking consistently outperforms no masking across all datasets, validating the effectiveness of our proposed Masked Client-Specific Learning approach. Third, despite the optimal mask ratio varying slightly among datasets, a mask ratio of 0.6 provides the best trade-off, maximizing test accuracy and ensuring robust performance across different datasets. This balance confirms that a mask ratio of 0.6 is optimal for enhancing model performance and stability in diverse scenarios. Based on these findings, we have chosen 0.6 as the mask ratio for all subsequent experiments.
\begin{figure}[ht]
    \centering
\includegraphics[width=0.5\textwidth]{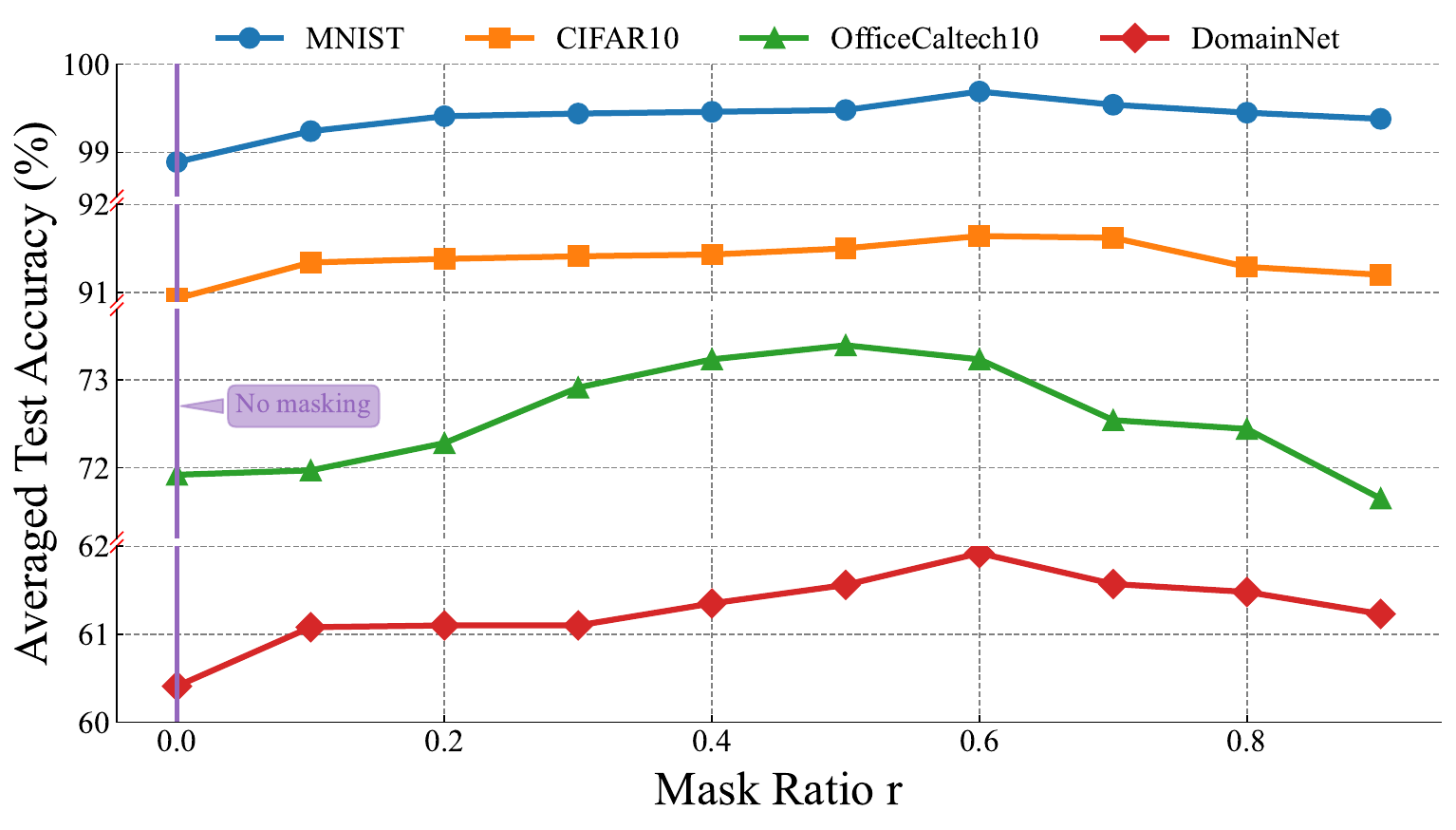}
    \caption{The impact of different mask ratios on the performance of the FedRIR method.} 
    \Description{}
    \label{fig:mask_ratio}
\end{figure}

\subsubsection{Effect of Heterogeneity Degree $\alpha$}
We conduct experiments on the FMNIST dataset under varying degrees of heterogeneity by adjusting the parameter $\alpha$ in the practical setting. The smaller the $\alpha$ is, the more likely the clients hold samples from one class. The averaged test accuracy for different values of $\alpha$ is presented in Table \ref{table_alpha}. The results show that as $\alpha$ increases, indicating less heterogeneity, the performance of FedAvg improves steadily, suggesting that FedAvg benefits from more homogeneous data distributions across clients. Conversely, all other pFL methods exhibit a decline in performance as $\alpha$ increases. This trend highlights their struggle with more homogeneous data distributions, where the need for strong personalization is reduced. Notably, FedRIR shows significant improvement over other methods, demonstrating superior robustness across all degrees of heterogeneity. As $\alpha$ increases, the performance gap between FedRIR and the second-best method widens, confirming the efficacy of FedRIR in handling varying degrees of data heterogeneity.

\begin{table}[ht]
\setlength{\tabcolsep}{2pt}  
\centering
\caption{The averaged test accuracy (\%) of the FMNIST classification for heterogeneity.}
\begin{tabular}{l|cccc}
\hline \toprule
              & $\alpha=0.1$     & $\alpha=0.3$    & $\alpha=0.5$    & $\alpha=1.0$    \\  \midrule
FedAvg        & 71.99$\pm$0.24   & 77.41$\pm$0.01 & 79.16$\pm$0.12 & 79.56$\pm$0.06 \\
Per-FedAvg    & 93.19$\pm$0.12   & 86.46$\pm$0.05 & 84.07$\pm$0.03 & 82.94$\pm$0.03 \\
FedProto      & 92.19$\pm$0.04   & 85.24$\pm$0.07 & 80.50$\pm$0.06 & 79.20$\pm$0.00 \\
Ditto         & \underline{95.98$\pm$0.01}   & \underline{90.07$\pm$0.02} & 86.90$\pm$0.04 & \underline{85.20$\pm$0.02} \\
FedRep        & 95.47$\pm$0.03   & 89.40$\pm$0.05 & 86.89$\pm$0.03 & 84.71$\pm$0.05 \\
FedRoD        & 95.57$\pm$0.02   & 89.49$\pm$0.15 & \underline{87.39$\pm$0.13} & 85.14$\pm$0.19 \\
FedBABU       & 95.21$\pm$0.13   & 87.95$\pm$0.57 & 85.72$\pm$0.01 & 83.55$\pm$0.31 \\
FedALA        & 95.56$\pm$0.00   & 89.52$\pm$0.11 & 86.10$\pm$0.10 & 84.90$\pm$0.08 \\
FedKD         & 95.17$\pm$0.03   & 89.29$\pm$0.03 & 85.92$\pm$0.06 & 84.60$\pm$0.06 \\
FedCP         & 95.19$\pm$0.09   & 88.75$\pm$0.06 & 85.47$\pm$0.09 & 84.13$\pm$0.08 \\ \midrule
\textbf{FedRIR} & \textbf{97.51$\pm$0.00} & \textbf{92.53$\pm$0.02} & \textbf{90.72$\pm$0.03} & \textbf{89.66$\pm$0.03} \\  \midrule
$\Delta$ \text{SOTA} & $\uparrow$ 1.53 & $\uparrow$ 2.46 & $\uparrow$ 3.33 & $\uparrow$ 4.46  \\ \bottomrule
\end{tabular}
\label{table_alpha}
\end{table}

\subsection{Ablation Study}
\begin{figure}
    \centering
    \includegraphics[width=\columnwidth]{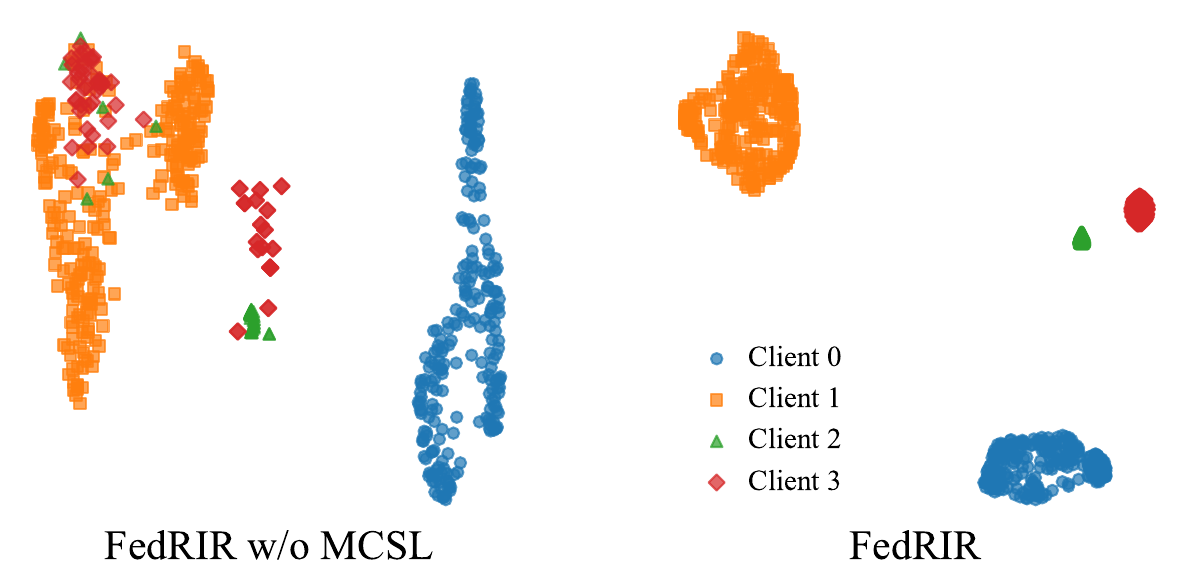}
    \caption{t-SNE visualization of client-specific features extracted by FedRIR w/o MCSL and FedRIR on OfficeCaltech10 dataset, with color represents a different client.}
    \Description{}
    \label{fig:OfficeCaltech10_specific}
\end{figure}

The ablation study in Table \ref{table_ablation_study} highlights the importance of the key components, Masked Client-Specific Learning (MCSL) and Information Distillation (ID)  in FedRIR, along with the impact of setting the masking ratio $r$ to zero. When the masking mechanism is disabled (FedRIR $r=0$), performance declines across all datasets, particularly on Cifar100, where accuracy drops from 72.07\% to 69.28\%, emphasizing that masking helps prevent overfitting and improves feature extraction. Removing MCSL (FedRIR w/o MCSL) results in an even larger performance decrease, particularly on Cifar100 (to 68.88\%) and OfficeCaltech10 (to 71.44\%), demonstrating that client-specific feature learning is essential for handling diverse data distributions. Interestingly, on OfficeCaltech10, which represents a real-world dataset with significantly heterogeneous client data, the absence of ID (FedRIR w/o ID) yields slightly better results (72.38\%) compared to removing MCSL. This suggests that in real-world datasets like OfficeCaltech10, where client data distributions vary widely, MCSL plays a more crucial role in achieving personalization, as capturing the unique characteristics of each client’s data is more important than refining global features. Thus, while ID is essential for aligning global knowledge, in datasets with high client-level variability, such as real-world scenarios, MCSL is the key driver of performance by tailoring the model to each client's specific data, which is evident in Figure \ref{fig:OfficeCaltech10_specific}. Without MCSL (left), client-specific features are poorly separated, leading to significant overlap between client data. In contrast, FedRIR (right) shows clear clustering of client features, highlighting the importance of MCSL in handling diverse data distributions effectively.

\begin{table}[ht]
\setlength{\tabcolsep}{2pt}  
    \centering
    \caption{Ablation Study}
    \begin{tabular}{l|ccc}
    \hline \toprule
    \textbf{Settings}   &  \textbf{Pathological} & \textbf{Practical} & \textbf{\makecell[c]{Real-world}} \\ \midrule
    Datasets            &  Cifar100 & Cifar10 & OfficeCaltech10 \\ \midrule
    FedRIR              & \textbf{72.07$\pm$0.12} & \textbf{91.64$\pm$0.03} & \textbf{73.23$\pm$0.59} \\ 
    FedRIR ($r=0$)      & 69.28$\pm$0.04 & 90.93$\pm$0.02 & 71.92$\pm$0.56 \\ 
    FedRIR w/o MCSL      & 68.88$\pm$0.03 & 90.76$\pm$0.10 & 71.44$\pm$0.48 \\ 
    FedRIR w/o ID       & 68.37$\pm$0.06 & 90.68$\pm$0.14 & 72.38$\pm$1.46 \\ \bottomrule
    \end{tabular}
    \label{table_ablation_study}
\end{table}

\section{Conclusion}
In this paper, we introduced FedRIR, a novel framework for federated learning that balance personalization and generalization through Masked Client-Specific Learning (MCSL) and Information Distillation (ID). Extensive experiments across pathological, practical, and real-world scenarios demonstrate that FedRIR consistently outperforms state-of-the-art methods in feature representation and classification accuracy. The key innovation lies in client-specific representation learning and refining global shared information, enhancing personalization and generalization. FedRIR also shows remarkable scalability and stability with varying client participation and comparable communication overhead, making it a robust and efficient solution for federated learning.

\bibliographystyle{ACM-Reference-Format}
\bibliography{sample-base}

\end{document}